\documentclass[letterpaper, 10 pt, conference]{ieeeconf}  
\usepackage{amsmath,amsfonts}
\usepackage{algorithmic}
\usepackage{array}
\usepackage[caption=false,font=normalsize,labelfont=sf,textfont=sf]{subfig}
\usepackage{textcomp}
\usepackage{stfloats}
\usepackage{url}
\usepackage{verbatim}
\usepackage{graphicx}

\usepackage{textcomp}
\usepackage{epstopdf}
\usepackage{epsfig}
\usepackage{soul}
\usepackage{multirow}
\usepackage{bbding}
\usepackage{paralist}
\usepackage[table,xcdraw]{xcolor}
\usepackage{booktabs}
\usepackage{colortbl}
\usepackage{adjustbox}

\usepackage{makecell}
\usepackage[section]{placeins}
\usepackage{hyperref}
\hypersetup{hypertex=true,
            colorlinks=true,
            linkcolor=blue,
            anchorcolor=blue,
            citecolor=blue}

\def\BibTeX{{\rm B\kern-.05em{\sc i\kern-.025em b}\kern-.08em
    T\kern-.1667em\lower.7ex\hbox{E}\kern-.125emX}}
\usepackage{balance}

\definecolor{rowgray1}{gray}{1.0}
\definecolor{rowgray2}{gray}{0.92}
\definecolor{rowgray3}{gray}{0.83}
\definecolor{mygreen}{RGB}{0,173,0}
\newcommand{\smallpm}{\text{\scriptsize$\pm$}}

\newcommand{\true}{\textcolor{blue}{\checkmark}}
\newcommand{\false}{\textcolor{red}{\texttimes}}

\IEEEoverridecommandlockouts                              

\title{\LARGE \bf
Differential Amplifier-Inspired AmpAttention for Multi-View Robotic Manipulation
}

\author{Jin Yang, Ping Wei$^*$, and Nanning Zheng
\thanks{National Key Laboratory of Human-Machine Hybrid
Augmented Intelligence, and Institute of Artificial Intelligence and Robotics, Xi'an Jiaotong University}
\thanks{* Corresponding author. pingwei@xjtu.edu.cn}
}

\begin{document}

\maketitle
\thispagestyle{empty}
\pagestyle{empty}

\begin{abstract}
Multi-view robotic manipulation methods with the attention mechanism have recently achieved significant progress in both training efficiency and task performance. However, the inherent redundancy, occlusion, and viewpoint dependency in robotic view images often lead to severe attention drift. To address this challenge, we propose AmpAttention, a novel attention mechanism inspired by differential amplifiers in analog circuits. It aims to suppress attention noise and capture high signal-to-noise ratio signals for more reliable perception. Based on this, we introduce the RVAF model, which integrates task-guided intra-view and inter-view AmpAttention. Compared to previous state-of-the-art methods, RVAF achieves the optimal average success rate across 18 RLBench tasks (249 variations) while reducing training time by 33.3\%. RVAF also demonstrates strong potential in real-world high-precision tasks, exemplified by its ability to pick up a dart and accurately insert it into the red bullseye. Furthermore, we extend RVAF to RVAF++ by incorporating the SAM2 image encoder. RVAF++ achieves substantial gains on high-precision tasks, achieving a 91\% success rate on the `insert peg' task. More qualitative results are provided at the anonymous project website \href{https://anonymous.4open.science/w/RVAF-Anonymization}{https://anonymous.4open.science/w/RVAF-Anonymization}.
\end{abstract}

\section{INTRODUCTION}
Robotic manipulation in unstructured 3D environments requires both precise reasoning and adaptability. Multi-view based methods have recently emerged as a powerful paradigm, as they provide diverse visual cues that are critical for accurate 3D understanding and task execution \cite{shridhar2023peract,goyal2023rvt,goyal2024rvt2,fang2025sam2act}. Compared with voxel-based representations, view-based methods such as RVT \cite{goyal2023rvt} and RVT-2 \cite{goyal2024rvt2} achieve superior performance with dramatically reduced training costs, demonstrating their potential for scalable deployment.

\begin{figure}[t]
  \centering
  \includegraphics[width=0.9\linewidth]{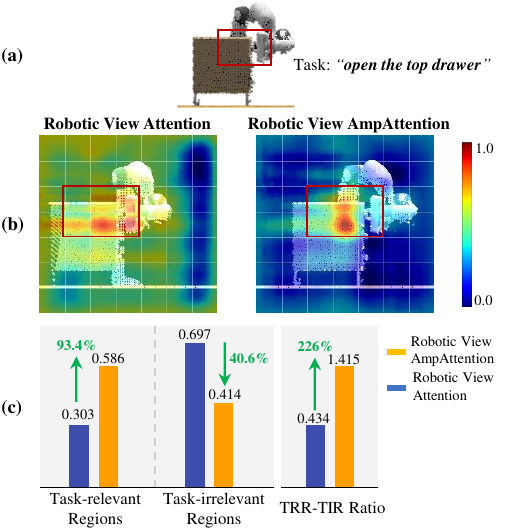}
  \caption{Comparison of attention distributions between standard attention and AmpAttention in robotic view images. (a) Task scenario of \textit{open the top drawer}. (b) Visualization of attention distributions under different attention mechanisms. left: standard attention. right: AmpAttention. (c) Quantitative comparison of both attention mechanisms.} 
  \label{fig:introduction}
\end{figure}

However, effectively exploiting multi-view observations remains challenging. Existing approaches typically rely on Transformer-based attention mechanisms \cite{goyal2023rvt,goyal2024rvt2,zhang2024same} to localize task-relevant regions (\textit{e.g.} the red box in Fig. \ref{fig:introduction}.(a)). But they often suffer from \textit{attention drift}, where focus shifts from meaningful cues to redundant or noisy content (\textit{e.g.}, background clutter, occlusions, or repeated patterns), as illustrated in the heatmap on the left of Fig. \ref{fig:introduction}.(b). This arises because robotic views contain highly redundant and viewpoint-dependent information, while standard attention has limited capacity to suppress irrelevant signals, ultimately degrading reasoning and action accuracy.

Given these challenges, there is a clear motivation to develop a reliable attention mechanism that can effectively capture task-relevant cues while suppressing task-irrelevant or noisy cues. To this end, we propose AmpAttention, a novel attention mechanism inspired by the differential amplifier in analog circuits. It focuses on the differential components of the signal while suppressing common-mode noise. This is achieved by jointly modeling the differential and common-mode components and optimizing attention learning through a Common Mode Rejection Ratio (CMRR) loss. By maintaining a high signal-to-noise ratio, AmpAttention ensures robust performance across multiple manipulation tasks.

As illustrated on the right of Fig. \ref{fig:introduction}.(b), AmpAttention produces more concentrated and discriminative activations around the manipulation target. Quantitative analysis in Fig. \ref{fig:introduction}.(c) shows that AmpAttention achieves a 93.4\% relative improvement in attention allocation to task-relevant regions (TRR) while reducing attention to task-irrelevant regions (TIR) by 40.6\%. This results in a 226\% relative improvement in the TRR–TIR ratio, significantly enhancing the model's ability to focus on the most relevant cues.

Building on AmpAttention, we propose the RVAF (Robotic View AmpFormer) model. It integrates intra-view AmpAttention to highlight task cues within each view, and inter-view AmpAttention to aggregate complementary information across views. Furthermore, we develop RVAF++, which leverages the SAM2 image encoder \cite{ravi2024sam2} to inject rich visual priors from large-scale pretraining \cite{fang2025sam2act,zhang2024same}.

Extensive experiments in both the RLBench simulation (18 RLBench tasks, 249 variations) and real-world settings demonstrate the effectiveness of our approach. RVAF achieves higher success rates than RVT-2 with 33\% less training time, while RVAF++ delivers remarkable gains on high-precision tasks such as `insert peg' (91\%). Extensive ablation studies validate the effectiveness of the components in the model. We further show that RVAF/RVAF++ outperforms the mainstream VLA baselines on some tasks. In real-world scenarios, we evaluate RVAF on five tasks using only a single third-person view camera. With only 50 collected demonstrations per task, RVAF performs well on high-precision manipulation tasks. In addition, RVAF demonstrates strong generalizability and robustness to environmental variations, maintaining stable performance under unseen objects and changing lighting conditions.


We summarize our \textbf{four key contributions} as follows.
\begin{compactitem}
\item [1.] We propose AmpAttention, a differential-amplifier-inspired mechanism that extracts high signal-to-noise task cues from robotic views.
\item [2.] We design RVAF, which integrates intra- and inter-view AmpAttention for multi-view manipulation.
\item [3.] We extend RVAF to RVAF++ by incorporating the SAM2 image encoder, significantly improving high-precision manipulation performance.
\item [4.] We achieve optimal results in both simulation and the real world, demonstrating task generalization, efficiency, and scalability.
\end{compactitem}

\section{Related Work}
\subsection{Vision-based Robotic Manipulation.}
Vision-based robotic manipulation policies have gained significant attention for their ability to provide high-dimensional state information. Researchers have explored various visual inputs and training paradigms for vision-based policies. Some methods \cite{yarats2021mastering} encode RGB-D observations into latent states for reinforcement learning, but training remained sample-inefficient and slow \cite{james2022qattention}. Vision-language-action models \cite{li2024cogact,kim2024openvla,wen2025tinyvla} integrate RGB observations with natural language instructions for end-to-end action prediction. While improving generalization, they typically rely on large-scale pretraining, making them resource-intensive. Some methods \cite{huang2024fourier,shridhar2023peract} voxelize point clouds to obtain strong structural priors. However, the precision of voxelization directly impacts task accuracy. Higher precision voxels require more memory and longer training times. 

An alternative line of work leverages multi-view representations to balance efficiency and accuracy. Instead of operating directly on point clouds, RVT \cite{goyal2023rvt} and RVT-2 \cite{goyal2024rvt2} re-render multiple virtual view images from reconstructed point clouds and extracts scene features with the Transformer. Our method follows this multi-view paradigm but focuses on enhancing training efficiency and robust extraction of task-relevant cues with a high signal-to-noise ratio. Additionally, with the advancement of visual foundation models, methods like SAM-E \cite{zhang2024same} and SAM2Act \cite{fang2025sam2act} have demonstrated that large-scale vision models provide richer visual representations, further enhancing model performance.

\subsection{Transformers for Robotic Manipulation.}

Transformers have been widely used in robotic manipulation, e.g., 3D scene representation \cite{gervet2023act3d}, intent prediction \cite{clever2021assistive}, and long-horizon control \cite{dasari2024dit,team2024octo}. However, many of these methods still require hundreds of real-world demonstrations, which limits their scalability and practicality.

Attention is central to Transformers, enabling modeling of dependencies and fusion of heterogeneous inputs, which has driven multimodal robot learning \cite{zhao2025vlas,li2024manipllm,singh2023progprompt,guhur2023instruction}. However, standard attention often suffers from \textit{attention drift} in vision-based manipulation (Fig.~\ref{fig:introduction}), diverting focus from task-relevant to irrelevant cues and reducing stability. This limitation motivates our design of a new attention mechanism to improve robustness.


\begin{figure}[htb]
  \centering
  \includegraphics[width=0.5\linewidth]{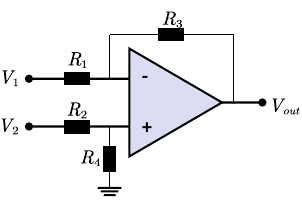}
  \caption{Illustration of the differential amplifier.}
  \label{fig:diffamp}
\end{figure}

\section{Methodology}
\subsection{Preliminaries}
\textbf{Differential Amplifier.}
A differential amplifier amplifies the difference between two input signals (differential component) while suppressing their identical part (common-mode component), which is typically noise. Fig. \ref{fig:diffamp} shows the circuit diagram of a differential amplifier. When $R_1=R_2,R_3=R_4$, the ideal output (zero common-mode gain) is given by
\begin{equation}
\label{eq:vout_diff}
    V_{out}=A_d\left( V_2-V_1 \right),   
\end{equation}
where $A_d$ is the differential gain. In practical circuits, component mismatches make common-mode signals unavoidable \cite{sansen2007analog}, leading to
\begin{equation}
\label{eq:vout_fact}
    V_{out}=A_dV_d+A_cV_c=A_d\left( V_2-V_1 \right) +\frac{A_c}{2}\left( V_1+V_2 \right),  
\end{equation}
where $A_c$ is the common-mode gain and $\frac{A_d}{A_c}$ refers to the common-mode rejection ratio (CMRR). A higher CMRR indicates that the circuit preserves more useful information while effectively rejecting interference.


\subsection{Overview} 
\textbf{Problem Statement.} We aim to train a robotic manipulation model capable of handling various tasks. The model takes as input a natural language description of the task along with the current visual observation and the current gripper state. The output is the next action, represented by the 6-DoF end-effector pose (3-DoF for position translation and 3-DoF for rotation orientation), a 1-DoF gripper state (open or close), and a 1-DoF collision state (a binary flag indicating whether collision is permissible during motion planning \cite{shridhar2023peract}). 

\begin{figure}[!tp]
  \centering
  \includegraphics[width=1.0\linewidth]{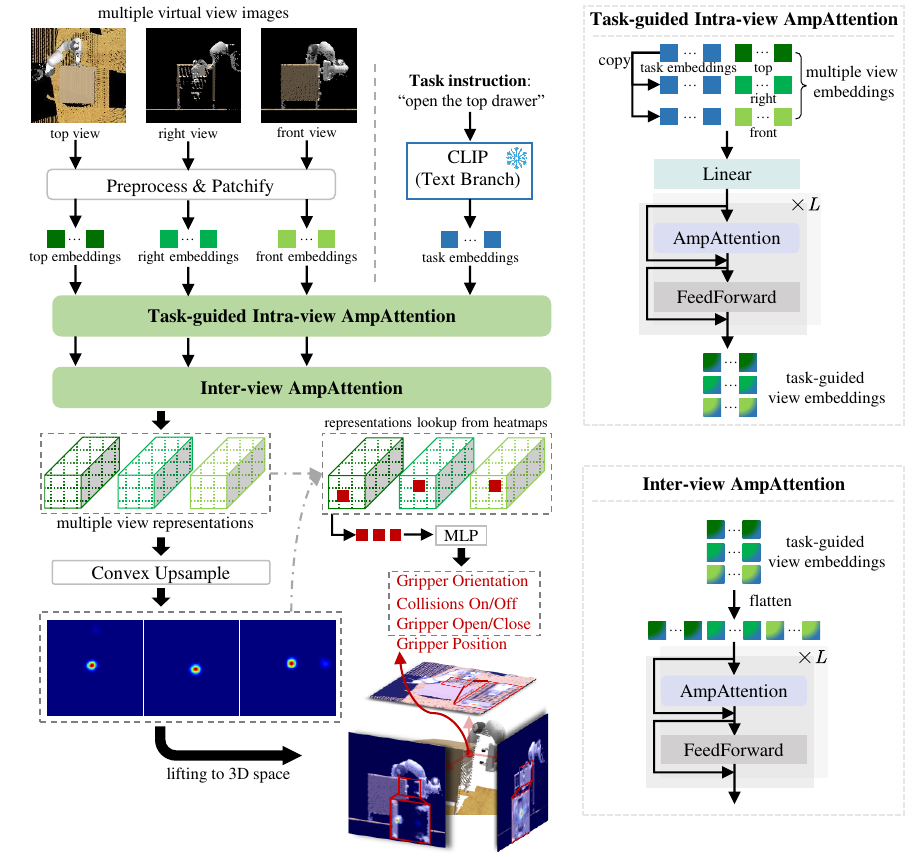}  
  \caption{Overview of the RVAF architecture. Given a task instruction and multi-view images, RVAF first encodes them into task embeddings and view-specific embeddings. These embeddings are then processed through intra-view and inter-view AmpAttention to capture view representations. The resulting representations are used to predict the next key-frame pose.}
  \label{fig:pipeline}
\end{figure}

We construct a dataset $D=\left\{ D_1,D_2,...,D_m \right\} $ comprising $m$ expert demonstrations across various tasks for model training. Each demonstration $D_i=\left( O_i,R_i,l_i \right) $ represents a successful trajectory of length $t_i$. It consists of a sequence of RGB-D observations $O_i=\left\{ o_{1}^{i},o_{2}^{i},...,o_{t_i}^{i} \right\} $, paired with corresponding robot actions $R_i=\left\{ r_{1}^{i},r_{2}^{i},...,r_{t_i}^{i} \right\}$, and a task instruction $l_i$ in natural language. 

We adopt a key-frame based paradigm for manipulation learning \cite{shridhar2023peract}, where key-frames denote critical gripper motions corresponding to meaningful state transitions. Following RVT \cite{goyal2023rvt}, we extract key-frames from raw demonstrations to construct training samples.

\textbf{Architecture Overview.} 
We illustrate the architecture of RVAF in Fig. \ref{fig:pipeline}. The input consists of multi-view orthographic RGB images, which are preprocessed and patchified into view-specific embeddings. Task instructions in natural language are encoded into task embeddings using the pre-trained CLIP \cite{radford2021clip}. These embeddings are first processed by the proposed intra-view AmpAttention, which guides each view to focus on task-relevant cues. Inter-view AmpAttention then aggregates complementary cues across views to form enriched view representations. The view representations are first processed with a convex upsampling layer to produce per-view heatmaps, which are back-projected into 3D space and aggregated to determine the target gripper position. Following RVT-2, we extract local features from each view based on the predicted position. These features are concatenated and processed by the MLP to predict the gripper orientation, collision state, and gripper state.

\begin{figure}[t]
  \centering
  \includegraphics[width=0.9\linewidth]{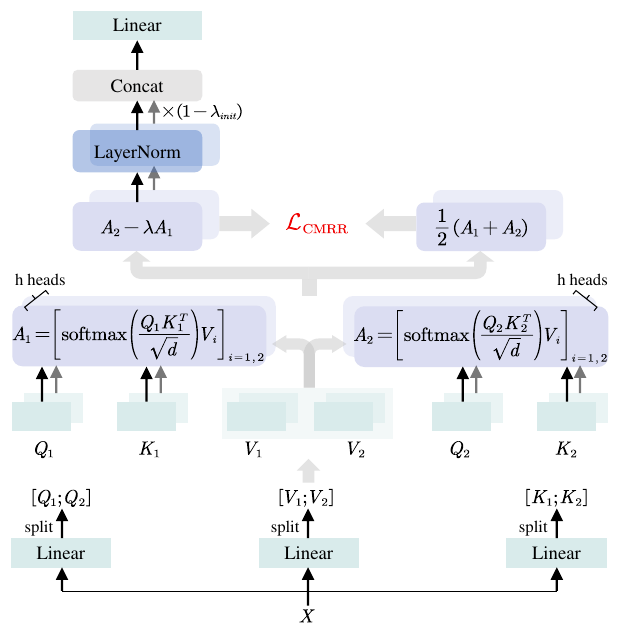}  
  \caption{Overview of the AmpAttention.}
  \label{fig:ampattention}
\end{figure}

\subsection{AmpAttention}
Existing view-based robotic manipulation methods rely on standard attention to extract task-relevant cues \cite{goyal2023rvt,zhang2024same}. However, the complexity and variability of manipulation scenarios present significant challenges in attention allocation. For example, in scenes with multiple visually similar objects, distractors can easily misdirect attention to irrelevant regions. Under such conditions, the standard attention mechanism is prone to attention drift (Fig.~\ref{fig:introduction}) due to its limited noise suppression capacity. To address this, we propose AmpAttention. It is designed to suppress noise and emphasize task-relevant cues with a higher signal-to-noise ratio.

The computational pipeline of AmpAttention is depicted in Fig. \ref{fig:ampattention}. Given an input sequence embedding $X\in\mathbb{R}^{N_\text{seq} \times d}$, it is first projected into the \textit{query}, \textit{key}, and \textit{value} through three separate `Linear' layers. Each projected embedding is then evenly split into two groups, resulting in $Q_1,Q_2,K_1,K_2,V_1,V_2\in\mathbb{R}^{N_\text{seq} \times \frac{d}{2}}$.
\begin{equation}
\label{eq:qkv}
    \left[ Q_1;Q_2 \right] =XW^Q, \left[ K_1;K_2 \right] =XW^K, \left[ V_1;V_2 \right] =XW^V,   
\end{equation}
where $W^Q,W^K,W^V\in\mathbb{R}^{d\times d}$ are trainable parameters. Then, we compute the attention output for each group as follows.
\begin{equation}
\label{eq:attn_diff_cm_tmp}
    \begin{split}   
    A_1\left( X \right) \,\,&=\,\,\left[ \mathrm{soft}\max\mathrm{(}\frac{Q_1K_{1}^{T}}{\sqrt{d_{Q_1}}})V_i \right] _{i=1,2},\\ 
    A_2\left( X \right) \,\,&=\,\,\left[ \mathrm{soft}\max\mathrm{(}\frac{Q_2K_{2}^{T}}{\sqrt{d_{Q_2}}})V_i \right] _{i=1,2},
    \end{split}    
\end{equation}
where $\left[ \cdot \right] _{i=1,2}$ denotes the concatenation of the results computed for indices $i=1,2$. The dimensions $d_{Q_1}$ and $d_{Q_2}$ correspond to those of $Q_1$ and $Q_2$, respectively. Moreover, the softmax attention formulation in Equation \ref{eq:attn_diff_cm_tmp} enables us to leverage efficient attention computation frameworks, such as FlashAttention \cite{dao2023flashattention2}, thereby improving training efficiency.
Inspired by Equations \ref{eq:vout_diff} and \ref{eq:vout_fact}, we define the differential attention operator $A_{\mathrm{diff}}$ as the difference between two softmax attention outputs, and the common-mode attention operator $A_{\mathrm{cm}}$ as the average of these two outputs. 
\begin{equation}
\label{eq:attn_diff_cm}
    \begin{split}   
    A_{\mathrm{diff}}\left( X \right) &=A_2-\lambda A_1, \\
    A_{\mathrm{cm}}\left( X \right) &=\frac{1}{2}\left( A_1+A_2 \right),
    \end{split}    
\end{equation}
where $A_1$ and $A_2$ represent the softmax attention values computed from the $Q$, $K$, and $V$ for each group. $\lambda$ is a learnable coefficient that balances the two attention branches in the differential computation, parameterized following \cite{ye2024differentialtransformer}.


As analyzed in Equation \ref{eq:vout_diff}, $A_{\mathrm{diff}}$ can directly serve as the output of AmpAttention in the ideal case. However, due to the inevitable existence of common-mode signals, Equation \ref{eq:vout_fact} motivates a more principled formulation of AmpAttention. One straightforward solution is to use a weighted sum of $A_{\mathrm{diff}}$ and $A_{\mathrm{cm}}$ as the final AmpAttention output. This design introduces additional hyperparameters that could potentially undermine model stability and generalization. To mitigate these issues, we propose a CMRR-inspired loss function $L_\text{CMRR}$, which optimizes differential attention learning. To avoid numerical instability arising from the division by zero during training, we approximate the CMRR by subtracting logarithms instead of directly performing the division. $L_\text{CMRR}$ is computed as follows.
\begin{equation}
\label{eq:cmrr_loss}
    \begin{split}   
    \mathcal{L}_{\mathrm{CMRR}}&=\frac{1}{N_{\mathrm{seq}}}\sum_{n=1}^{N_{\mathrm{seq}}}{S\left( A_{\mathrm{cm}}^{n},A_{\mathrm{diff}}^{n} \right)},\\
    S\left( a,b \right) &=\max \left( 0,\log \left( 1+\left\| a \right\| _2 \right) -\log \left( 1+\left\| b \right\| _2 \right) \right) ,
    \end{split}   
\end{equation}
where $\left\| \cdot \right\| _2$ denotes the Frobenius norm. The hinge operation with $\max(0, \cdot)$ ensures that the loss imposes an effective constraint only when common-mode noise suppression is insufficient. This design prevents unnecessary regularization from disrupting well-learned differential characteristics. $S\left( A_{\mathrm{cm}}^{n},A_{\mathrm{diff}}^{n} \right)$ refers to the CMRR loss value at the $n$-th position in the input sequence. The $\mathcal{L}_{\mathrm{CMRR}}$ computes the mean of these CMRR loss values across all positions in the sequence of length $N_{\mathrm{seq}}$.

We also adopt the multi-head attention (MHA) design where each head’s output is individually normalized and scaled. The processed outputs from all heads are then concatenated to produce the final results as follows.
\begin{equation}
\label{eq:ampattention_mh}
    \begin{split}   
    &\mathrm{head}_h=A_{\mathrm{diff}}^{h}\left( X;W_{h}^{Q},W_{h}^{K},W_{h}^{V}\right), s.t. \,\mathcal{L}_{\mathrm{CMRR}}\left( A_{\mathrm{diff}}^{h},A_{\mathrm{cm}}^{h} \right)
    \\
    &\mathrm{head}_{h}^{\prime}=\left( 1-\lambda _{\mathrm{init}} \right) \cdot \mathrm{LN}\left( \mathrm{head}_h \right) \,\, \\
    &\mathrm{AmpAttention}\left( X \right) =\mathrm{Concat}\left[ \mathrm{head}_{h}^{\prime} \right] _{h=1,2,...,H}W^O
    \end{split}    
\end{equation}
where $\lambda _{\mathrm{init}}\in(0,1)$ initializes $\lambda$ with a strategy consistent with \cite{ye2024differentialtransformer}. $\mathrm{LN}\left( \cdot \right)$ is the layer normalization operation, we adopt RMSNorm for each head in this work. $\left( 1-\lambda _{\mathrm{init}} \right)$ 
scales the normalized output of each head \cite{ye2024differentialtransformer}, ensuring that the gradient dynamics of AmpAttention approximate those of standard MHA. This design preserves training stability and enables direct hyperparameter transfer. $H$ is the number of heads. $\mathrm{Concat}\left[ \cdot \right]$ refers to concatenate all heads along the channel dimension. $W^O\in \mathbb{R}^{d\times d}$ denotes the learnable output projection. We set $H$ to half the number of heads in standard MHA, ensuring that the head dimension of AmpAttention matches that of standard MHA. This configuration maintains computational complexity comparable to standard MHA while avoiding performance degradation.

\subsection{Task-guided Intra-view AmpAttention}
A key aspect of multi-view-based robotic manipulation is the extraction of task-relevant visual cues from multiple views. Prior methods apply self-attention within views and then fuse with task instructions, which can cause early focus on irrelevant regions. We instead adopt a ``task-guided first'' design. We first apply cross-AmpAttention between task instructions and each view to guide feature extraction, and then use inter-view AmpAttention to aggregate visual cues.

The computation process of Task-guided Intra-view AmpAttention is illustrated in the upper-right part of Fig. \ref{fig:pipeline}. The task embeddings are first concatenated with each view embedding. These concatenated embeddings are then fed into a Linear layer for mapping, which helps eliminate the modality gap. The mapped features are subsequently processed through a stack of $L$ modules ($L$ is set to 4), resulting in task-guided view embeddings. Each module contains AmpAttention and FeedForward submodules, with residual connections incorporated to stabilize training. The FeedForward submodule consists of two Linear layers, with a GEGLU \cite{shazeer2020glu} activation function applied between them.

\subsection{Inter-view AmpAttention}
The Inter-view AmpAttention plays a critical role in integrating task-relevant cues across multiple views. Its computation process is illustrated in the lower-right part of Fig. \ref{fig:pipeline}. Specifically, it begins with task-guided view embeddings, which are flattened into a single sequence. These flattened embeddings are then processed through a stack of modules containing both AmpAttention and FeedForward submodules. The configuration of these modules is identical to that of the Intra-view AmpAttention, consisting of $L$ modules.

For action prediction, we follow RVT-2 \cite{goyal2024rvt2}, which employs convex upsampling for position estimation, location-conditioned rotation prediction, and simple classifiers for gripper and collision states.

\subsection{RVAF++: SAM2-Enhanced Visual Representation}
Previous works \cite{fang2025sam2act,zhang2024same} have demonstrated that utilizing visual foundation models pre-trained on large-scale datasets to extract visual embeddings significantly improves the performance of view-based robotic manipulation methods. To this end, we integrate SAM2's image encoder \cite{ravi2024sam2} into RVAF to extract view image embeddings, resulting in RVAF++. For model efficiency, we employ low-rank adaptation \cite{hu2022lora} with a default rank of 16 to fine-tune the image encoder of SAM2. 

\subsection{Training}
To train the model, we define the model loss function $\mathcal{L} _{\mathrm{model}}$, which consists of $\mathcal{L} _{\mathrm{CMRR-total}}$, $\mathcal{L} _{\mathrm{trans}}$, $\mathcal{L} _{\mathrm{rota}}$, $\mathcal{L} _{\mathrm{gripper}}$, and $\mathcal{L} _{\mathrm{collision}}$. The total CMRR loss $\mathcal{L} _{\mathrm{CMRR-total}}$ is employed to regulate the learning of all AmpAttention modules within the model, ensuring the provision of high signal-to-noise ratio visual cues. It is computed as the average of the CMRR losses across all AmpAttention modules, and further averaged across $N$ samples.
\begin{equation}
\label{eq:cmrr_total_loss}
\mathcal{L} _{\mathrm{CMRR-total}}=\,\, \frac{1}{N}\sum_{i=1}^N{\left( \frac{1}{2L}\sum_{l=1}^{2L}{\mathcal{L}_{\mathrm{CMRR}}^{l}}\left( A_{\mathrm{diff}}^{i},A_{\mathrm{cm}}^{i} \right) \right)} 
\end{equation}
where $2L$ is the total number of AmpAttention modules in the model (Fig. \ref{fig:pipeline}). The CMRR loss for each AmpAttention module is computed as described in Equation \ref{eq:cmrr_loss}.

We follow prior works \cite{goyal2023rvt,goyal2024rvt2} and apply standard cross-entropy losses for action prediction, including translation loss $\mathcal{L} _{\mathrm{trans}}$, rotation loss $\mathcal{L} _{\mathrm{rota}}$, gripper state loss $\mathcal{L} _{\mathrm{gripper}}$, and collision state loss $\mathcal{L} _{\mathrm{collision}}$. Thus, the model loss function $\mathcal{L} _{\mathrm{model}}$ is defined as follows.
\begin{equation}
\label{eq:model_loss}
\begin{aligned}
    \mathcal{L} _{\mathrm{model}}=&\alpha _1\left( \mathcal{L} _{\mathrm{trans}}+\mathcal{L} _{\mathrm{rota}}+\mathcal{L} _{\mathrm{gripper}}+\mathcal{L} _{\mathrm{collision}} \right) \\ & + \alpha _2\mathcal{L} _{\mathrm{CMRR}-\mathrm{total}} ,
\end{aligned}
\end{equation}
where $\alpha _1$ and $\alpha _2$ are balance hyperparameters. In this work, we set $\alpha _1$ to 1.0 and $\alpha _2$ to 0.01, respectively.


\begin{table*}[htbp]
    \centering
    \renewcommand{\arraystretch}{1.} 
    \caption{Multi-task performance (\%) on RLBench.}
    \label{tab:comparison}
    \begin{adjustbox}{max width=\textwidth}
        \begin{tabular}{lcccccccccc}
        \toprule
        \multicolumn{1}{l|}{Models} & \makecell*[c]{Avg. \\Success$\uparrow$} & \multicolumn{1}{c|}{\makecell*[c]{Train time\\(days)$\downarrow$}} & \makecell*[c]{Close \\ Jar} & \makecell*[c]{Drag \\ Stick} & \makecell*[c]{Insert \\ Peg} & \makecell*[c]{Meat off \\ Grill} & \makecell*[c]{Open \\ Drawer} & \makecell*[c]{Place \\ Cups}  & \makecell*[c]{Place \\ Wine} & \makecell*[c]{Push \\ Buttons} \\
        \midrule
        
        
        
        
        
        

        \rowcolor{rowgray1}
        \multicolumn{1}{l|}{ACt3D \cite{gervet2023act3d}}& 65.0 & \multicolumn{1}{c|}{5.0(V100)} & 92.0 & 92.0 & 27.0 & 94.0 & 93.0 & 3.0 & 80.0 & 99.0\\
        
        \rowcolor{rowgray2}
        \multicolumn{1}{l|}{RVT \cite{goyal2023rvt}}& 62.9 & \multicolumn{1}{c|}{1.0(V100)}& 
        52.0{\smallpm}\scriptsize{2.5} & 
        99.2{\smallpm}\scriptsize{1.6} & 
        11.2{\smallpm}\scriptsize{3.0} & 
        88.0{\smallpm}\scriptsize{2.5} & 
        71.2{\smallpm}\scriptsize{6.9} & 
        4.0{\smallpm}\scriptsize{2.5} & 
        91.0{\smallpm}\scriptsize{5.2} & 
        \textbf{100.0}{\smallpm}\scriptsize{0.0} \\
        
        \rowcolor{rowgray1}
        \multicolumn{1}{l|}{RVT-2 \cite{goyal2024rvt2}} & 81.4 & \multicolumn{1}{c|}{0.83(V100)} & 
        \textbf{100.0}{\smallpm}\scriptsize{0.0} & 
        99.0{\smallpm}\scriptsize{1.7} & 
        40.0{\smallpm}\scriptsize{0.0} & 
        99.0{\smallpm}\scriptsize{1.7} & 
        74.0{\smallpm}\scriptsize{11.8} & 
        38.0{\smallpm}\scriptsize{4.5} & 
        95.0{\smallpm}\scriptsize{3.3} & 
        \textbf{100.0}{\smallpm}\scriptsize{0.0} \\

        \rowcolor{rowgray2}
        \multicolumn{1}{l|}{RVT-2$^*$ \cite{goyal2024rvt2}} & 80.3 & \multicolumn{1}{c|}{0.36(4090)} & 
        \textbf{100.0}{\smallpm}\scriptsize{0.0} & 
        \textbf{100.0}{\smallpm}\scriptsize{0.0} & 
        38.0{\smallpm}\scriptsize{5.2} & 
        98.0{\smallpm}\scriptsize{4.0} & 
        81.0{\smallpm}\scriptsize{3.8} & 
        36.0{\smallpm}\scriptsize{3.3} & 
        94.0{\smallpm}\scriptsize{2.3} & 
        92.0{\smallpm}\scriptsize{0.0} \\
        
        \rowcolor{rowgray1}
        \multicolumn{1}{l|}{SAM-E \cite{zhang2024same}}& 70.6 & \multicolumn{1}{c|}{-} & 
        82.4{\smallpm}\scriptsize{3.6} & 
        \textbf{100.0}{\smallpm}\scriptsize{0.0} & 
        18.4{\smallpm}\scriptsize{4.6} & 
        95.2{\smallpm}\scriptsize{3.3} & 
        \textbf{95.2}{\smallpm}\scriptsize{5.2} & 
        0.0{\smallpm}\scriptsize{0.0} & 
        94.4{\smallpm}\scriptsize{4.6} & 
        \textbf{100.0}{\smallpm}\scriptsize{0.0} \\

        \rowcolor{rowgray2}
        \multicolumn{1}{l|}{SAM2Act \cite{fang2025sam2act}} & 86.8 & \multicolumn{1}{c|}{-(H100)}& 99.0{\smallpm}\scriptsize{2.0} & 99.0{\smallpm}\scriptsize{2.0} & 84.0{\smallpm}\scriptsize{5.7} & 98.0{\smallpm}\scriptsize{2.3} & 83.0{\smallpm}\scriptsize{6.0} & \textbf{47.0}{\smallpm}\scriptsize{6.0} & 93.0{\smallpm}\scriptsize{3.8} & \textbf{100.0}{\smallpm}\scriptsize{0.0}\\

        \rowcolor{blue!10}
        \multicolumn{1}{l|}{\textbf{RVAF} (ours)} & \textbf{83.1} & \multicolumn{1}{c|}{\textbf{0.24}(4090)}& 
        \textbf{100.0}{\smallpm}\scriptsize{0.0} & 
        \textbf{100.0}{\smallpm}\scriptsize{0.0} & 
        30.0{\smallpm}\scriptsize{4.0} & 
        \textbf{100.0}{\smallpm}\scriptsize{0.0} & 
        86.0{\smallpm}\scriptsize{2.3} & 
        31.0{\smallpm}\scriptsize{3.8} & 
        \textbf{96.0}{\smallpm}\scriptsize{3.3} & 
        \textbf{100.0}{\smallpm}\scriptsize{0.0} \\

        \rowcolor{blue!10}
        \multicolumn{1}{l|}{\textbf{RVAF++} (w. SAM2)} & \textbf{87.0} & \multicolumn{1}{c|}{0.58(4090)}& 
        99.0{\smallpm}\scriptsize{2.0} & 
        \textbf{100.0}{\smallpm}\scriptsize{0.0} & 
        \textbf{91.0}{\smallpm}\scriptsize{3.8} & 
        \textbf{100.0}{\smallpm}\scriptsize{0.0} & 
        85.0{\smallpm}\scriptsize{2.0} & 
        36.0{\smallpm}\scriptsize{3.3} & 
        95.0{\smallpm}\scriptsize{3.8} & 
        \textbf{100.0}{\smallpm}\scriptsize{0.0} \\

        \midrule
        \multicolumn{1}{l|}{Models} & \makecell*[c]{Put in \\ Cupboard} & \makecell*[c]{Put in \\ Drawer} & \makecell*[c]{Put in \\ Safe} & \makecell*[c]{Screw \\ Bulb} & \makecell*[c]{Slide \\ Block} & \makecell*[c]{Sort \\ Shape} & \makecell*[c]{Stack \\ Blocks} & \makecell*[c]{Stack \\ Cups} & \makecell*[c]{Sweep to \\ Dustpan} & \makecell*[c]{Turn \\ Tap} \\
        \midrule
        
        
        

        \rowcolor{rowgray1}
        \multicolumn{1}{l|}{ACt3D \cite{gervet2023act3d}}& 51.0 & 90.0 & 95.0 & 47.0 & 93.0 & 8.0 & 12.0 & 9.0 & 92.0 & 94.0\\
        
        \rowcolor{rowgray2}
        \multicolumn{1}{l|}{RVT \cite{goyal2023rvt}}& 
        49.6{\smallpm}\scriptsize{3.2} & 
        88.0{\smallpm}\scriptsize{5.7} & 
        91.2{\smallpm}\scriptsize{3.0} & 
        48.0{\smallpm}\scriptsize{5.7} & 
        81.6{\smallpm}\scriptsize{5.4} & 
        36.0{\smallpm}\scriptsize{2.5} & 
        28.8{\smallpm}\scriptsize{3.9} & 
        26.4{\smallpm}\scriptsize{8.2} &
        72.0{\smallpm}\scriptsize{0.0} & 
        93.6{\smallpm}\scriptsize{4.1} \\
        
        \rowcolor{rowgray1}
        \multicolumn{1}{l|}{RVT-2 \cite{goyal2024rvt2}} & 66.0{\smallpm}\scriptsize{4.5} & 
        96.0{\smallpm}\scriptsize{0.0} & 
        96.0{\smallpm}\scriptsize{2.8} & 
        88.0{\smallpm}\scriptsize{4.9} & 
        92.0{\smallpm}\scriptsize{2.8} & 
        35.0{\smallpm}\scriptsize{7.1} & 
        80.0{\smallpm}\scriptsize{2.8} & 
        69.0{\smallpm}\scriptsize{5.9} & 
        \textbf{100.0}{\smallpm}\scriptsize{0.0} & 
        99.0{\smallpm}\scriptsize{1.7}\\

        \rowcolor{rowgray2}
        \multicolumn{1}{l|}{RVT-2$^*$ \cite{goyal2024rvt2}} & 68.0{\smallpm}\scriptsize{3.3} & 
        99.0{\smallpm}\scriptsize{2.0} & 
        94.0{\smallpm}\scriptsize{4.0} & 
        92.0{\smallpm}\scriptsize{0.0} & 
        51.0{\smallpm}\scriptsize{3.8} & 
        49.0{\smallpm}\scriptsize{7.6} & 
        \textbf{83.0}{\smallpm}\scriptsize{3.8} & 
        78.0{\smallpm}\scriptsize{2.3} & 
        99.0{\smallpm}\scriptsize{2.0} & 
        93.0{\smallpm}\scriptsize{5.0}\\
        
        \rowcolor{rowgray1}
        \multicolumn{1}{l|}{SAM-E \cite{zhang2024same}}& 
        64.0{\smallpm}\scriptsize{2.8} & 
        92.0{\smallpm}\scriptsize{5.7} & 
        95.0{\smallpm}\scriptsize{3.3} & 
        78.4{\smallpm}\scriptsize{3.6} & 
        \textbf{95.2}{\smallpm}\scriptsize{1.8} & 
        34.4{\smallpm}\scriptsize{6.1} & 
        26.4{\smallpm}\scriptsize{4.6} & 
        0.0{\smallpm}\scriptsize{0.0} & 
        \textbf{100.0}{\smallpm}\scriptsize{0.0} & 
        \textbf{100.0}{\smallpm}\scriptsize{0.0}\\

        \rowcolor{rowgray2}
        \multicolumn{1}{l|}{SAM2Act \cite{fang2025sam2act}} & 75.0{\smallpm}\scriptsize{3.8} & 99.0{\smallpm}\scriptsize{2.0} & 98.0{\smallpm}\scriptsize{2.3} & 89.0{\smallpm}\scriptsize{2.0} & 86.0{\smallpm}\scriptsize{4.0} & \textbf{64.0}{\smallpm}\scriptsize{4.6} & 76.0{\smallpm}\scriptsize{8.6} & 78.0{\smallpm}\scriptsize{4.0} &
        99.0{\smallpm}\scriptsize{2.0} &
        96.0{\smallpm}\scriptsize{5.7} \\

        \rowcolor{blue!10}
        \multicolumn{1}{l|}{\textbf{RVAF} (ours)} &
        \textbf{76.0}{\smallpm}\scriptsize{0.0} & 
        \textbf{100.0}{\smallpm}\scriptsize{0.0} & 
        94.0{\smallpm}\scriptsize{2.3} & 
        93.0{\smallpm}\scriptsize{2.0} & 
        80.0{\smallpm}\scriptsize{0.0} & 
        47.0{\smallpm}\scriptsize{2.0} & 
        72.0{\smallpm}\scriptsize{3.3} & 
        \textbf{91.0}{\smallpm}\scriptsize{3.8} & 
        \textbf{100.0}{\smallpm}\scriptsize{0.0} & 
        99.0{\smallpm}\scriptsize{2.0}\\

        \rowcolor{blue!10}
        \multicolumn{1}{l|}{\textbf{RVAF++} (w. SAM2)} &
        73.0{\smallpm}\scriptsize{7.6} & 
        \textbf{100.0}{\smallpm}\scriptsize{0.0} & 
        \textbf{99.0}{\smallpm}\scriptsize{2.0} & 
        \textbf{95.0}{\smallpm}\scriptsize{2.0} & 
        74.0{\smallpm}\scriptsize{2.3} & 
        61.0{\smallpm}\scriptsize{7.6} & 
        76.0{\smallpm}\scriptsize{3.3} & 
        84.0{\smallpm}\scriptsize{4.6} & 
        99.0{\smallpm}\scriptsize{2.0} & 
        99.0{\smallpm}\scriptsize{2.0}\\

        \bottomrule
        \end{tabular}
    \end{adjustbox}
    
\end{table*}

\section{Experiments}
\label{sec:section_4}

\subsection{Simulation Experiments Setup}
\textbf{Simulation Benchmark.} Following prior works~\cite{goyal2023rvt,goyal2024rvt2,fang2025sam2act}, we evaluate on RLBench~\cite{james2020rlbench}, a simulation suite built on CoppeliaSim with a Franka Panda arm and parallel gripper. The robot is equipped with four RGB-D cameras (front, left shoulder, right shoulder, and wrist) at a resolution of 128×128. RLBench provides 18 manipulation tasks with 249 variations, ranging from simple actions to common pick-and-place and high-precision operations. 

\textbf{Training and Evaluation Details.} Following existing baselines~\cite{goyal2024rvt2,zhang2024same}, we utilize 1800 expert demonstrations (100 demonstrations per task) from the RLBench~\cite{james2020rlbench} for training and 450 unseen demonstrations (25 demonstrations per task) for evaluation. The resolution of all virtual images is set to 224 × 224. All models are trained using 8 NVIDIA 4090 (24 GB) GPUs. We train the RVAF model for approximately 50K steps, with a batch size of 160 (20 \text{$\times$} 8) and a learning rate of 2e-3. The RVAF++ model is trained for approximately 110K steps, with a batch size of 64 (8 \text{$\times$} 8) and a learning rate of 3.2e-3. Both RVAF and RVAF++ utilize the LAMB~\cite{you2019lamb} optimizer and employ cosine learning rate decay with a warm-up period for the first 2,000 steps. 

We evaluate all tasks using the model from the final epoch. Due to the randomness introduced by the sampling-based motion planner used in RLBench~\cite{james2020rlbench}, we run each model four times on each task and report the average success rate and standard deviation.


\begin{figure*}[!tp]
  \centering
  \includegraphics[width=1.\linewidth]{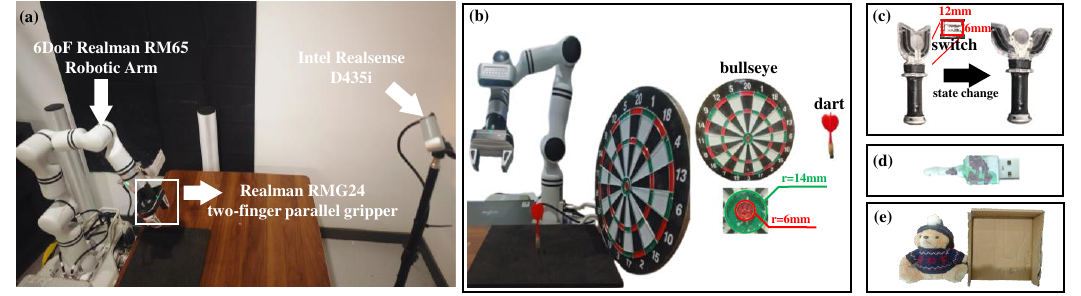}
  \caption{Real-world experimental setup. (a) Real-world task scenario setting. (b) Pick dart and insert bullseye: the red/green bullseye has a radius of 6/14 mm. (c) Press the toy switch: pressing a 12mm $\times$ 6mm control switch. (d) Cable grasping task. (e) Pick and place toy bear task.}
  \label{fig:real_world}
\end{figure*}

\subsection{Real-World Experiments Setup}
\textbf{Environment Setup and Tasks.} 
We construct a tabletop experimental platform as illustrated in Fig. \ref{fig:real_world}(a). The system consists of a Realman RM65 6DoF robotic arm equipped with the Realman RMG24 two-finger parallel gripper. The scene is perceived by a statically mounted Intel RealSense D435i camera from a third-person perspective. The extrinsic transformation between the camera and the robot base frame is obtained via a checkerboard-based calibration method. 

To evaluate precision and generalization, we design five real-robot tasks: four challenging tasks (`press the toy switch', `pick dart and insert red/green bullseye' and `cable grasping') and one easier `pick and place toy bear' task. Each task supports diverse natural language instructions and includes randomized object placements to test robustness. The task scene details are shown in Fig.~\ref{fig:real_world}(b)-(e).

\textbf{Data Collection.}
We collect paired visual and action data using a manual teaching device. The visual data consists of RGB and depth images captured by a single camera, while the action data includes the robot arm's joint states and the gripper status. The visual data is recorded at 20 Hz and action data at 200 Hz. We align the two data streams using their timestamps to ensure synchronization. For each task, we collected 50 demonstration sequences.

\textbf{Training and Evaluation Details.} 
To ensure a fair comparison, we train both the RVT-2 \cite{goyal2024rvt2} and RVAF models on real-world data using the same training settings. For different tasks, we select different model checkpoints saved during training. During evaluation, we deploy the models on a server using the Flask framework to create an HTTP-based inference API. Observation data collected on a local device is sent to the server via POST requests. Upon receiving the data, the server performs inference using the deployed model and returns the predictions in real time. 

\subsection{Quantitative Experiments Results}

\textbf{Simulation Multi-task Performance.} Tab. \ref{tab:comparison} compares the performance of our RVAF-based models with previous state-of-the-art methods. To ensure fairness, we re-train RVT-2 \cite{goyal2024rvt2}, using our hardware setup and refer to this result as RVT-2$^*$. Without incorporating vision foundation models, RVAF achieves a 3.5\% relative improvement over RVT-2 on the same device, while also reducing training time by roughly 33.3\% (from 0.36 days to 0.24 days). When leveraging SAM2, RVAF++ surpasses the prior method SAM-E \cite{zhang2024same} by 16\% absolute improvement. Overall, RVAF-based models outperform previous methods in 12 out of 18 tasks.

Further analysis indicates significant performance improvements in high-precision tasks after the introduction of the SAM2 \cite{ravi2024sam2}. For example, in the \textit{insert peg} task, success rates increased substantially from 84\% (prior best) to 91\%. We attribute these improvements to the powerful representational capabilities provided by SAM2, which enables more high-precision visual cues. However, RVAF++ does not achieve performance comparable to previous best methods on the \textit{slide blocks} task. We argue that this discrepancy is due to overfitting on the simple task or inherent trade-offs in multi-task learning, where improvements on complex tasks negatively impact performance on simpler ones.

\begin{table*}[htbp]
    \centering
    \renewcommand{\arraystretch}{0.9} 
    \caption{Ablation study (\%) on RLBench. We quantify the impact of different components in RVAF. }
    \label{tab:ablation}
    \begin{adjustbox}{max width=\textwidth}
        \begin{tabular}{lccccccccccc}
        \toprule
        \multicolumn{1}{c|}{\makecell*[c]{Row \\ ID}} & 
        \makecell*[c]{AmpAttn.} & 
        \makecell*[c]{AmpAttn. \\ w/o $\mathcal{L}_{\mathrm{CMRR}}$} & 
        \makecell*[c]{Attn.} & 
        \makecell*[c]{Intra-Inter} & 
        \makecell*[c]{Coarse-to- \\ Fine} &
        \makecell*[c]{Scale of \\ Zoom in} &
        \makecell*[c]{$\#$ of \\ Views} &
        \makecell*[c]{Train time \\ (hours)} &
        \makecell*[c]{Train time \\ ($\%$ of base)} &
        \makecell*[c]{Avg. \\ Success} &
        \makecell*[c]{Avg. Success\\ diff. wrt. base} \\
        \midrule
        \rowcolor{rowgray1}
        \multicolumn{1}{c|}{1} & 
        \true & \false & \false & \true & \true & 
        4 & 3 & 5.876 & 100.0$\%$ & 83.1 & 0 \\

        \rowcolor{rowgray1}
        \multicolumn{1}{c|}{2} & 
        \false & \true & \false & \true & \true & 
        4 & 3 &  5.849 &  99.5$\%$ & 81.6  &  -1.5 \\

        \rowcolor{rowgray1}
        \multicolumn{1}{c|}{3} & 
        \false & \false & \true & \true & \true & 
        4 & 3 &  5.150 &  87.6$\%$ &   81.2&   -1.9\\

        \rowcolor{rowgray2}
        \multicolumn{1}{c|}{4} & 
        \true & \false & \false & \false & \true & 
        4 & 3 & 5.870 &  99.9$\%$ &  80.2 &  -2.9 \\
        

        \rowcolor{rowgray1}
        \multicolumn{1}{c|}{5}& 
        \true & \false & \false & \true & \false & 
        - & 3 & 4.172  &  71.0$\%$ &  67.2 &  -15.9 \\
        
        \rowcolor{rowgray2}
        \multicolumn{1}{c|}{6}& 
        \true & \false & \false & \true & \true & 
        2 & 3 & 5.879  &  100.1$\%$ &  82.8 & -0.3 \\
        
        \rowcolor{rowgray2}
        \multicolumn{1}{c|}{7}& 
        \true & \false & \false & \true & \true & 
        3 & 3 & 5.868  &  99.9$\%$ &  81.8 & -1.3 \\
        
        \rowcolor{rowgray2}
        \multicolumn{1}{c|}{8}& 
        \true & \false & \false & \true & \true & 
        5 & 3 &  6.007 &  102.2$\%$ & 79.9 & -3.2 \\
        \rowcolor{rowgray1}
        \multicolumn{1}{c|}{9}& 
        \true & \false & \false & \true & \true & 
        4 & 2 &  5.238 &  89.1$\%$ &  76.2 &  -6.9 \\

        \rowcolor{rowgray1}
        \multicolumn{1}{c|}{10}& 
        \true & \false & \false & \true & \true & 
        4 & 4 &  7.315 &  124.5$\%$ &   80.7&   -2.4\\
   
        \bottomrule
        \end{tabular}
    \end{adjustbox}
    
\end{table*}

\textbf{Simulation Ablation Study.} Tab. \ref{tab:ablation} presents extensive ablation studies of components in RVAF within the simulation environment. We examine the impact of several design choices: a) The type of attention mechanism used in the RVAF model. `AmpAttn.' is the full AmpAttention mechanism, `AmpAttn. w/o $\mathcal{L}_{\mathrm{CMRR}}$' is a variant that considers only differential gain (as defined in Equation \ref{eq:vout_diff}), and `Attn.' refers to the standard attention mechanism; b) whether to apply task-guided intra-view AmpAttention before inter-view AmpAttention (`Intra-Inter'); c) whether to use the coarse-to-fine design \cite{james2022c2farm,goyal2024rvt2} (`Coarse-to-Fine'); d) varying the zoom-in scale used in the fine stage of coarse-to-fine rendering (`Scale of Zoom in'); e) the number of rendered virtual views (`$\#$ of Views'). For each setting, we report both the average success rate and training time.

(a) Comparing rows 1, 2, and 3, the RVAF model with the full AmpAttention mechanism achieves the optimal performance. This improvement is attributed to its ability to suppress attention noise, allowing the model to focus more effectively on high signal-to-noise visual cues. The performance drop from row 1 to row 2, where the $\mathcal{L}_{\mathrm{CMRR}}$ loss is removed, indicates that common-mode suppression contributes to more stable and precise attention.
(b) The comparison between rows 1 and 4 highlights the benefit of applying task-guided intra-view AmpAttention before inter-view AmpAttention, resulting in improved average success rate.
(c) As expected, adopting the coarse-to-fine design leads to significant gains in performance (row 1 vs. row 5). 
(d) The scale of zoom-in in the fine stage also plays a critical role. Comparing rows 1, 6, 7, and 8 reveals that appropriate scaling provides a more informative field of view, which helps to improve model performance. However, larger zoom levels do not guarantee better performance and even reduce efficiency by increasing training time and limiting contexts.
(e) The number of rendered virtual views is a critical design factor in multi-view models. Comparing rows 1, 9, and 10, we observe that increasing the number of views does not necessarily improve performance. Excessive views may introduce redundancy and amplify attention noise, leading to higher computational cost without performance gains.

\begin{table}[h]\small
    \centering
    \renewcommand{\arraystretch}{1.0}
    \setlength\tabcolsep{7pt}
    \caption{Comparison of RVAF series with VLAs on four tasks. Tasks 1 to 4 correspond to close jar, insert onto square peg, open drawer, and put item in drawer, respectively.}
    \label{table:compare_vlas}
    \scalebox{0.9}{
    \begin{tabular}{c|c c c c c}
    \toprule
    \multicolumn{1}{c|}{\multirow{1}{*}{Method}} &Task1 &Task2 &Task3 &Task4 & Avg.\\
    \midrule
    \multirow{1}{*}{RVAF} &100.0 &30.0 &86.0 &100.0 &79.0\\
    \multirow{1}{*}{RVAF++} &99.0 &91.0 &85.0 &100.0 &\textbf{93.8}\\
    \midrule
    \multirow{1}{*}{$\pi_0\text{-FAST}$\cite{black2024pi_0}} &62.1 &8.9 &75.0 &68.3 &53.6\\
    \multirow{1}{*}{OpenVLA-OFT\cite{kim2025openvlaoft}} &68.3 &12.3 &81.2 &70.1 &58.0\\
    \bottomrule
    \end{tabular}
    }
\end{table}

\begin{table*}[htbp]
    \centering
    \renewcommand{\arraystretch}{0.8} 
    \caption{Performance of the model in real-world scenarios.} 
    \label{tab:real_world}
    \begin{adjustbox}{max width=\textwidth}
        \begin{tabular}{lcccccc}  
        \toprule
      
        \multicolumn{1}{l|}{Models} & \multicolumn{1}{c|}{\makecell*[c]{Avg. Success$\uparrow$}(\%)} & \makecell*[c]{Press the \\toy switch} & \makecell*[c]{Pick dart and \\insert red bullseye} & \makecell*[c]{Pick dart and \\insert green bullseye} & \makecell*[c]{Cable grasping} & \makecell*[c]{Pick and place \\toy bear} \\
        \midrule
        
        \rowcolor{rowgray1}
        \multicolumn{1}{l|}{RVT-2 \cite{goyal2024rvt2}} &  \multicolumn{1}{c|}{44.0} & 3/10 & 3/10  & 5/10 & 2/10 & 9/10 \\

        \rowcolor{rowgray1}
        \multicolumn{1}{l|}{OpenVLA-OFT\cite{kim2025openvlaoft} } &  \multicolumn{1}{c|}{44.0} & 3/10 & 2/10 & 4/10  & 3/10 & \textbf{10/10}\\
                
        \rowcolor{rowgray1}
        \multicolumn{1}{l|}{RVAF (ours)}& \multicolumn{1}{c|}{\textbf{64.0}} & \textbf{6/10} & \textbf{5/10} & \textbf{8/10}  & \textbf{4/10} & 9/10 \\


        \bottomrule
        \end{tabular}
    \end{adjustbox}
    
\end{table*}

\begin{figure}[!tp]
  \centering
  \includegraphics[width=1.\linewidth]{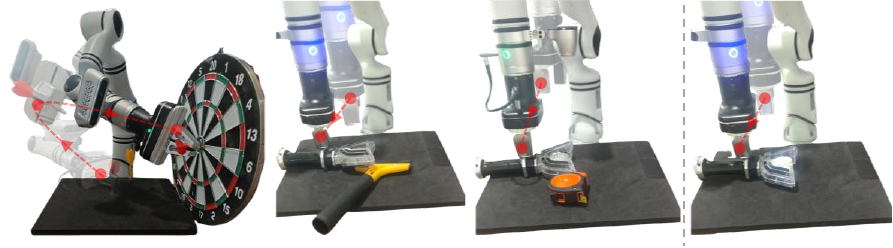}
  \caption{Robustness validation of RVAF under the real-world environment.}
  \label{fig:real_task_robo}
\end{figure}

\begin{figure}[t]
  \centering
  \includegraphics[width=0.93\linewidth]{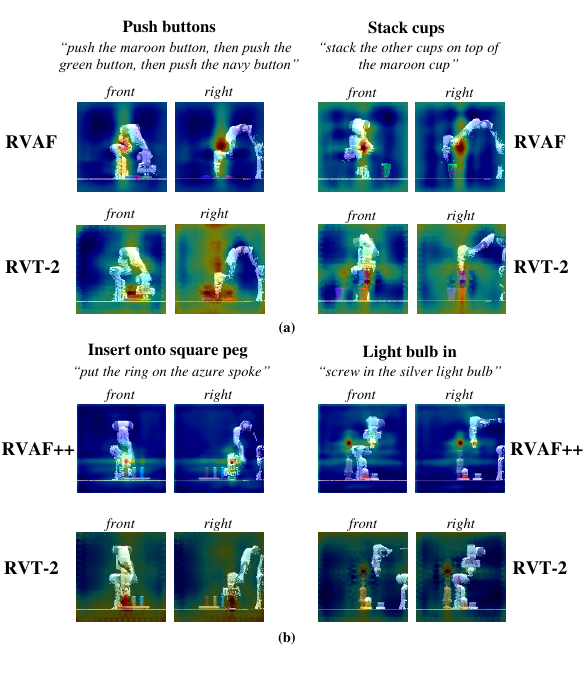} 
  \caption{Illustration of attention heatmaps for different models across various tasks. The highlighted regions indicate the robot’s next action-relevant area. (a) Basic manipulation tasks. (b) High-precision manipulation tasks. }
  \label{fig:heatmap_vis}
\end{figure}

\textbf{Compared to VLAs.} 
Tab.~\ref{table:compare_vlas} compares RVAF++ with two advanced VLAs ($\pi_0$-FAST and OpenVLA-OFT) on four tasks. RVAF++ achieves the best overall performance, with an average success rate of 93.8\%, substantially outperforming $\pi_0$-FAST and OpenVLA-OFT. Notably, on the high-precision assembly task (Task2), RVAF++ is significantly higher than both VLA models. These results suggest that the multi-view imitation learning paradigm still offers clear advantages on certain tasks.


\textbf{Real-World Performance.} 
Tab.~\ref{tab:real_world} compares RVAF with RVT-2 and OpenVLA-OFT on five real-world tasks (10 trials per task). RVAF consistently performs better on the four high-precision tasks, and achieves the highest average success rate (64.0\%).


Fig. \ref{fig:real_task_robo} validates the robustness of RVAF in real-world environments. The first three images show that the robotic arm successfully executes tasks despite interference from previously unseen objects. The last image demonstrates successful task execution under changing lighting conditions.


\subsection{Qualitative Results}
Fig. \ref{fig:heatmap_vis} compares attention heatmaps of RVAF, RVAF++, and RVT-2 on both basic tasks and high-precision tasks. The heatmaps from the front and right views reveal how each model allocates attention, offering intuitive insights into the behavioral differences observed in Tab.\ref{tab:comparison}.

RVT-2, based on standard attention, shows diffuse patterns vulnerable to background interference, leading to attention drift. RVAF with AmpAttention yields more concentrated focus, enabling accurate localization of task-relevant regions. RVAF++ further improves precision and semantic consistency in high-precision tasks, explaining its substantial performance gains in these scenarios.

\section{Conclusion}
In this work, we propose RVAF and RVAF++, which incorporate task-guided intra-view and inter-view AmpAttention to improve perception accuracy and training efficiency for multi-view robotic manipulation. At the core is AmpAttention, a novel attention mechanism inspired by differential amplifiers. It facilitates the extraction of task-relevant visual cues while suppressing irrelevant cues, thereby mitigating attention drift. Extensive simulation and real-world experiments demonstrate strong effectiveness, generalization, and robustness. Furthermore, our results confirm that leveraging large-scale pretrained visual foundation models significantly enhances the performance of robotic manipulation in high-precision tasks.



\bibliographystyle{IEEEtran}
\balance
\bibliography{ref}

\end{document}